\documentclass{article}

    \PassOptionsToPackage{square}{natbib}
    \PassOptionsToPackage{numbers}{natbib}
    \PassOptionsToPackage{sort}{natbib}

\usepackage[final]{neurips_2020}

\usepackage[utf8]{inputenc} 
\usepackage[T1]{fontenc}    
\usepackage[hidelinks]{hyperref}       
\usepackage{url}            
\usepackage{booktabs}       
\usepackage{amsfonts}       
\usepackage{nicefrac}       
\usepackage{microtype}      
\usepackage{multirow}
\usepackage[title]{appendix}
\usepackage[english]{babel}
\usepackage{graphicx}
\usepackage{multibib}

\newcites{SM}{SM References}

\usepackage{mkolar_definitions}
\usepackage{xcolor}
\usepackage{float}
\usepackage{caption}
\newcount\Comments  
\definecolor{darkgreen}{rgb}{0,0.5,0}
\definecolor{darkred}{rgb}{0.7,0,0}
\definecolor{teal}{rgb}{0.3,0.8,0.8}
\definecolor{blue}{rgb}{0,0,1}
\definecolor{purple}{rgb}{0.5,0,1}

\newcommand{\kibitz}[2]{\ifnum\Comments=1\textcolor{#1}{#2}\fi}

\makeatletter
\newcommand{\settitle}{\@maketitle}
\makeatother

\title{Model-Agnostic Graph Regularization \\ 
for Few-Shot Learning}

\author{
   Ethan Shen \\
   Department of Computer Science\\
   Stanford University \\
   \texttt{ezshen@cs.stanford.edu} \\
   \And
   Maria Brbi\'c \\
   Department of Computer Science \\
   Stanford University \\
   \texttt{mbrbic@cs.stanford.edu} \\
   \And
   Nicholas Monath \\
   College of Information and Computer Sciences \\
   University of Massachusetts Amherst \\
   \texttt{nmonath@cs.umass.edu} \\
   \And
   Jiaqi Zhai \\
   Google Research \\
   \texttt{jiaqi@jiaqizhai.com} \\
   \And
   Manzil Zaheer \\
   Google Research \\
   \texttt{manzilzaheer@google.com} \\
   \And
   Jure Leskovec \\
   Department of Computer Science \\
   Stanford University \\
   \texttt{jure@cs.stanford.edu} \\
}

\begin{document}

\maketitle

\newcommand{\dataset}{\ensuremath{X}}
\newcommand{\dataspace}{\ensuremath{\Xcal}}
\newcommand{\lbls}{\ensuremath{Y}}
\newcommand{\lblsspace}{\ensuremath{\Ycal}}
\begin{abstract}

In many domains, relationships between categories are encoded in the knowledge graph. Recently, promising results have been achieved by incorporating knowledge graph as side information in hard classification tasks with severely limited data. However, prior models consist of highly complex architectures with many sub-components that all seem to impact performance. In this paper, we present a comprehensive empirical study on graph embedded few-shot learning.  We introduce a graph regularization approach that allows a deeper understanding of the impact of incorporating graph information between labels. Our proposed regularization is widely applicable and model-agnostic, and boosts the performance of any few-shot learning model, including fine-tuning, metric-based and optimization-based meta-learning. Our approach improves performance of strong base learners by up to $2\%$ on Mini-ImageNet and $6.7\%$ on ImageNet-FS, outperforming state-of-the-art graph embedded methods. Additional analyses reveal that graph regularizing models results in a lower loss for more difficult tasks, such as those with fewer shots and less informative support examples.

\end{abstract}
\section{Introduction}

Few-shot learning refers to the task of generalizing from a very few examples, an ability that humans have but machines lack. Recently, major breakthroughs have been achieved with meta-learning, which leverages prior experience from many related tasks to effectively learn to adapt to unseen tasks ~\cite{schmidhuber87, bengio92}. At a high level, meta-learning has been divided into metric-based approaches that learn a transferable metric across tasks ~\cite{vinyals2016matching, snell2017prototypical, sung2018learning}, and optimization-based approaches that learn initializations for fast adaptation on new tasks ~\cite{finn2017model, rusu2018meta}. Beyond meta-learning, transfer learning by pretraining and fine-tuning on novel tasks has achieved surprisingly competitive performance on few-shot tasks ~\cite{chen2019closer, dhillon2019baseline, wang2019simpleshot}.

In many domains, external knowledge about the class labels can be used. For example, this information is crucial in the zero-shot learning paradigm, which seeks to generalize to novel classes without seeing any training examples ~\cite{kankuekul2012online, lampert2013attribute, xian2018zero}. Prior knowledge often takes the form of a knowledge graph ~\cite{wang2018zero}, such as the WordNet hierarchy ~\cite{miller1995wordnet} in computer vision tasks, or Gene Ontology ~\cite{ashburner2000gene} in biology. In such cases, relationships between categories in the graph are used to transfer knowledge from base to novel classes. This idea dates back to hierarchical classification ~\cite{koller1997hierarchically, salakhutdinov2011learning}. 

Recently, few-shot learning methods have been enhanced with graph information, achieving state-of-the-art performance on benchmark image classification tasks \cite{chen2019knowledge, liu2019prototype, liu2019learning, li2019large, suo2020tadanet}. Proposed methods typically employ sophisticated and highly parameterized graph models on top of convolutional feature extractors. However, the complexity of these methods prevents deeper understanding of the impact of incorporating graph information. Furthermore, these models are inflexible and incompatible with other approaches in the rapidly-improving field of meta-learning, demonstrating the need for a model-agnostic graph augmentation method.

Here, we conduct a comprehensive empirical study of incorporating knowledge graph information into few-shot learning. First, we introduce a \textit{graph regularization} approach for incorporating graph relationships between labels applicable to any few-shot learning method. Motivated by node embedding ~\cite{ grover2016node2vec} and graph regularization principles ~\cite{hallac2015network}, our proposed regularization enforces category-level representations to preserve neighborhood similarities in a graph. By design, it allows us to directly measure benefits of enhancing few-shot learners with graph information. We incorporate our proposed regularization into three major approaches of few-shot learning: (i) metric-learning, represented by Prototypical Networks ~\cite{snell2017prototypical}, (ii) optimization-based learning, represented by LEO ~\cite{rusu2018meta}, and (iii) fine-tuning, represented by SGM ~\cite{qiao2018few} and $\text{S2M2}_R$ ~\cite{mangla2020charting}. We demonstrate that graph regularization consistently improves each method and can be widely applied whenever category relations are available. Next, we compare our approach to state-of-the-art methods, including those that utilize the same category hierarchy on standard benchmark Mini-ImageNet and large-scale ImageNet-FS datasets. Remarkably, we find that our approach improves the performance of strong base learners by as much as $6.7\%$ and outperforms graph embedded baselines, even though it is simple, easy to tune, and introduces minimal additional parameters. Finally, we explore the behavior of incorporating graph information in controlled synthetic experiments. Our analysis shows that graph regularizing models yields better decision boundaries in lower-shot learning, and achieves significantly higher gains on more difficult few-shot episodes. 
\section{Model-Agnostic Graph Regularization}

Our approach is a model-agnostic graph regularization objective based on the idea that the graph structure of class labels can guide learning of model parameters. The graph regularization objective ensures labels in the same graph neighborhood have similar parameters. The regularization is combined with a classification loss to form the overall objective. The classification loss is flexible and depends on the base learner. For instance, the classification loss can correspond to cross-entropy loss  \cite{chen2019closer}, or distance-based loss between example embeddings and class prototoypes \cite{snell2017prototypical}. 

\subsection{Problem Setup}

We assume that we are given a dataset defined as a pair of examples $\dataset \subseteq \dataspace$ with corresponding labels $\lbls \subseteq \lblsspace$. We say that point $\textbf{x}_i \in \dataset$ has the label $y_i \in \lbls$. For each episode, we learn from a support set $\mathcal{D}_s = \{ (\textbf{x}_1, y_1), (\textbf{x}_2, y_2), ... , (\textbf{x}_K, y_K) \}$ and evaluate on a held-out query set $\mathcal{D}_q = \{ (\textbf{x}_1^*, y_1^*), (\textbf{x}_2^*, y_2^*), ... , (\textbf{x}_T^*, y_T^*) \}, \mathcal{D}_q \cap \mathcal{D}_s = \emptyset$. For each dataset, we split all classes into $C_{train}$ and $C_{test}$, $C_{train} \cap C_{test} = \emptyset$. During evaluation, we sample the $N$ classes from a larger set of classes $C_{test}$, and sample $K$ examples from each class. During training, we use a disjoint set of classes $C_{train}$ to train the model. Non-episodic training approaches treat $C_{train}$ as a standard supervised learning problem, while episodic training approaches match the conditions on which the model is trained and evaluated by sampling episodes from $C_{train}$. More details on the problem setup can be found in Appendix \ref{appendix:prob_setup}. Additionally, we assume that there exists side information about the labels in the form of a graph $G(\lblsspace, E)$ where $\lblsspace$ is the set of all nodes in the label graph, and $E$ is the set of edges.  

\subsection{Regularization}
We incorporate graph information using the random walk-based node2vec objective \citep{grover2016node2vec}. Random walk methods for graph embedding \citep{perozzi2014deepwalk} are fit by maximizing the probability of predicting the neighborhoods for each target node in the graph. Node2vec performs biased random walks by introducing hyperparameters to balance between breadth-first search (BFS) and depth-first search (DFS) to capture local structures and global communities. We formulate the node2vec loss below:
\begin{equation}
\label{eqn:reg}
\mathcal{L}_{graph}(G, \theta) = - \sum_{y \in \lblsspace} \bigg[ - \log Z_y + \sum_{n \in N(y)} \frac{1}{T} sim(\theta_{n}, \theta_y) \bigg],
\end{equation}
where $\theta$ are node representations, $sim$ is a similarity function between the nodes, $N(y)$ is the set of neighbor nodes of node $y$, $T$ is the temperature hyperparameter, and $Z_y$ is partition function defined as $Z_y = \sum_{v \in \lblsspace} \exp(\frac{1}{T}sim(\theta_y, \theta_v))$. The partition function is approximated using negative sampling \cite{mikolov2013distributed}. We obtain the neighborhood $N(y)$ by performing a random walk starting from a source node $y$. The similarity function $sim$ depends on the base learner, which we outline in Section \ref{section:aug}.

\subsection{Augmentation Strategies}
\label{section:aug}

Our graph-regularization framework is model-agnostic and intuitively applicable to a wide variety of few-shot approaches. Here, we describe augmentation strategies for high-performing learners from metric-based meta-learning, optimization-based meta-learning and fine-tuning by formulating each as a joint learning objective.

\subsubsection{Augmenting Metric-Based Models}
Metric-based approaches learn an embedding function to compare query set examples. Prototypical networks are a high-performing learner of this class, especially when controlling for model complexity \citep{chen2019closer, triantafillou2019meta}. Prototypical networks construct a prototype $p_j$ of the $j^{th}$ class by taking the mean of support set examples, and comparing query examples using Euclidean distance. We regularize these prototypes so they respect class similarities and get the joint objective:
\begin{equation}
    \sum_{(x_i, y_i) \in \mathcal{D}_s} \bigg[ \norm{x_i - p_{y_i}}^2_2 + \sum_{\substack{y' \in \lblsspace}} \exp(-\norm{x_i - p_{y'}}^2_2) \bigg] + \lambda \mathcal{L}_{graph}(G, \theta).
\end{equation}

We set the graph similarity function to negative Euclidean distance, $sim(p_i, p_j) = -\norm{p_i - p_j}^2_2$. Note that our approach can easily be extended to other metric-based learners, for example regularizing the output of the relation module for Relation Networks \citep{sung2018learning}.

\subsubsection{Augmenting Optimization-Based Models}
Optimization-based meta-learners such as MAML \citep{finn2017model} and LEO \citep{rusu2018meta} consist of two optimization loops: the outer loop updates the neural network parameters to an initialization that enables fast adaptation, while the inner loop performs a few gradient updates over the support set to adapt to the new task. Graph regularization enforces class similarities among parameters during inner-loop adaptation. 

Specifically for LEO, we pass support set examples through an encoder to produce latent class encodings $z$, which are decoded to generate classifier parameters $\theta$. Given instantiated model parameters learned from the outer loop, gradient steps are taken in the latent space to get $z'$ while freezing all other parameters to produce final adapted parameters $\theta'$. For more details, please refer to \cite{rusu2018meta}. Concretely, we obtain the joint regularized objective below for the inner-loop adaptations:
\begin{equation}
    \label{eqn:class_reg}
   \sum_{(x_i, y_i) \in \mathcal{D}_s} \bigg[ - z_{y_i}^T x_i + \sum_{y' \in \lblsspace} \exp(z_{y_i}^T x_i) \bigg] + \lambda \mathcal{L}_{graph}(G, z).
\end{equation}

We set the graph similarity function to the inner product, $sim(z_i, z_j) = z_i^T z_j$, though in practice cosine similarity, $sim(z_i, z_j) = z_i^T z_j / \norm{z_i}\norm{z_j}$ results in more stable learning.  

\subsubsection{Augmenting Fine-tuning Models}
Recent approaches such as Baseline++ \cite{chen2019closer} and $\text{S2M2}_R$ \cite{mangla2020charting} have demonstrated remarkable performance by pre-training a model on the training set, and fine-tuning the classifier parameters $\theta$ on the support set of each task. We follow \cite{chen2019closer} and freeze the feature embedding model during fine-tuning, though the model can be fine-tuned as well \cite{dhillon2019baseline}. We perform graph regularization on the classifiers in the last layer of the network, which are learned for novel classes during fine-tuning. This results in the objective below:
\begin{equation}
    \label{eqn:class_reg}
   \sum_{(x_i, y_i) \in \mathcal{D}_s} \bigg[ - \frac{x_i^T\theta_{y_i}}{\norm{x_i}\norm{\theta_{y_i}}}  + \sum_{y' \in \lblsspace} \exp\bigg(\frac{x_i^T\theta_{y_i}}{\norm{x_i}\norm{\theta_{y_i}}}\bigg) \bigg] + \lambda \mathcal{L}_{graph}(G, \theta).
\end{equation}

We set the graph similarity to cosine similarity, $sim(\theta_i, \theta_j) = \theta_i^T \theta_j / \norm{\theta_i}\norm{\theta_j}$.
\section{Experimental Results}

For all ImageNet experiments, we use the associated WordNet \cite{miller1995wordnet} category hierarchy to define graph relationships between classes. Details of the experimental setup are given in Appendix \ref{appendix:exp_setup}. On the synthetic dataset, we analyze the effect of graph regularizing few-shot methods. 

\subsection{Mini-ImageNet Experiments}
\label{section:mini_exps}

We compare performance to few-shot baselines and graph embedded approach KGTN \cite{chen2019knowledge} on the Mini-ImageNet experiment. We enhance  $\text{S2M2}_R$ \cite{mangla2020charting}, a strong baseline fine-tuning model. \autoref{tab:mini_imagenet_sota} shows graph regularization results on Mini-ImageNet compared to results of the state-of-the-art models. We find that $\text{S2M2}_R$ enhanced with the proposed graph regularization outperforms all other methods on both 1- and 5-shot tasks.

As an additional baseline, we consider KGTN which also utilizes the WordNet hierarchy for better generalization. To ensure that our improvements are not caused by the embedding function, we pretrain KGTN feature extractor using $\text{S2M2}_R$. Even when controlling for improvements in the feature extractor, we find that our simple graph regularization method outperforms complex graph-embedded models.

\begin{table}[H]
\centering
\caption{Results on $1$-shot and $5$-shot classification on the Mini-ImageNet dataset. We report average accuracy over 600 randomly sampled episodes. We show graph-based models in the bottom section.}
\begin{tabular}{p{5cm}p{2cm}p{2cm}p{2cm}}
\toprule
\bf Model & \bf Backbone & \bf 1-shot & \bf 5-shot \\
\midrule
Qiao \cite{qiao2018few} & WRN 28-10 & 59.60 $\pm$ 0.41 &  73.74 $\pm$ 0.19\\
Baseline++ \cite{chen2019closer} & WRN 28-10 & 59.62 $\pm$ 0.81 & 78.80 $\pm$ 0.61 \\
LEO (train+val) \cite{rusu2018meta} & WRN 28-10 & 61.76 $\pm$ 0.08 & 77.59 $\pm$ 0.12 \\
ProtoNet \cite{snell2017prototypical} & WRN 28-10 & 62.60 $\pm$ 0.20 & 79.97 $\pm$ 0.14 \\
MatchingNet \cite{vinyals2016matching} & WRN 28-10 & 64.03 $\pm$ 0.20 & 76.32 $\pm$ 0.16 \\
$\text{S2M2}_R$ \cite{mangla2020charting} & WRN 28-10 & 64.93 $\pm$ 0.18 & 83.18 $\pm$ 0.11 \\
SimpleShot \cite{wang2019simpleshot} & WRN 28-10 & 65.87 $\pm$ 0.20 & 82.09 $\pm$ 0.14 \\
\midrule 
KGTN \cite{chen2019knowledge} & WRN 28-10 & 65.71 $\pm$ 0.75 & 81.07 $\pm$ 0.50 \\
\textbf{$\text{S2M2}_R$ + Graph (Ours)} & WRN 28-10 & \textbf{66.93} $\pm$ \textbf{0.65} & \textbf{83.35} $\pm$ \textbf{0.53} \\
\bottomrule
\end{tabular}
\label{tab:mini_imagenet_sota}
\end{table}

\subsection{Graph Regularization is Model-Agnostic}
\label{section:agnostic}
 
We augment ProtoNet \cite{snell2017prototypical}, LEO \cite{rusu2018meta}, and $\text{S2M2}_R$ \cite{mangla2020charting} approaches with graph regularization and evaluate effectiveness of our approach on the Mini-ImageNet dataset.
These few-shot learning models are fundamentally different and vary in both optimization and training procedures. For example, ProtoNet and LEO are both trained episodically, while $\text{S2M2}_R$ is trained non-episodically. However, the flexibility of our graph regularization loss allows us to easily extend each method. \autoref{tab:agnostic} shows the results of graph enhanced few-shot baselines. The results demonstrate that graph regularization consistently improves performance of few-shot baselines with larger gains in the $1$-shot setup.

\begin{table}[H]
\caption{Performance of graph-regularized few-shot baselines on the Mini-ImageNet dataset. We report average accuracy over 600 randomly sampled episodes.}
\centering
\begin{tabular}{p{5cm}p{2cm}p{2cm}p{2cm}}
\toprule
\bf Model & \bf Backbone & \bf 1-shot & \bf 5-shot \\
\midrule
ProtoNet \cite{snell2017prototypical} & ResNet-18 & 54.16 $\pm$ 0.82 & 73.68 $\pm$ 0.65 \\

\textbf{ProtoNet + Graph (Ours)} & ResNet-18 & \textbf{55.47 $\pm$ 0.73} & \textbf{74.56 $\pm$ 0.49} \\
\midrule
LEO (train) \cite{rusu2018meta} & WRN 28-10 & 58.22 $\pm$ 0.09 & 74.46 $\pm$ 0.19 \\
\textbf{LEO  + Graph (Ours)} & WRN 28-10 & \textbf{60.93 $\pm$ 0.19} & \textbf{76.33 $\pm$ 0.17} \\
\midrule
$\text{S2M2}_R$ \cite{mangla2020charting} & WRN 28-10 & 64.93 $\pm$ 0.18 & 83.18 $\pm$ 0.11 \\
\textbf{$\text{S2M2}_R$ + Graph (Ours)} & WRN 28-10 & \textbf{66.93} $\pm$ \textbf{0.65} & \textbf{83.35} $\pm$ \textbf{0.53} \\
\bottomrule
\end{tabular}
\label{tab:agnostic}
\end{table}

\subsection{Large-Scale Few-Shot Classification}

We next evaluate our graph regularization approach on the large-scale ImageNet-FS dataset, which includes 1000 classes. Notably, this task is more challenging because it requires choosing among all novel classes, an arguably more realistic evaluation procedure. We sample \textit{K} images per category, repeat the experiments $5$ times, and report mean accuracy with $95\%$ confidence intervals. Results demonstrate that our graph regularization method boosts performance of the SGM baseline \cite{hariharan2017low} by as much as $6.7\%$. Remarkably, augmenting SGM with graph regularization outperforms all few-shot baselines, as well as models that benefit from class semantic information and label hierarchy such as KTCH \cite{liu2019large} and KGTN \cite{chen2019knowledge}. We include further experimental details in Appendix \ref{appendix:exp_setup}, and explore further ablations to justify design choices in Appendix \ref{appendix:ablations}. 

\begin{table}[H]
\centering
\caption{Top-5 accuracy on the novel categories for the Imagenet-FS dataset. KTCH and KGTN are graph-based models. We report $95\%$ confidence intervals where provided. The $95\%$ confidence intervals for \cite{hariharan2017low, vinyals2016matching, snell2017prototypical, wang2018low} are on the order of $0.2\%$.}
\begin{tabular}{p{3.5cm}p{2cm}p{2cm}p{2cm}p{2cm}}
\toprule
\bf Model & \bf Backbone & \textbf{1-shot} & \textbf{2-shot} & \textbf{5-shot} \\
\midrule
SGM \cite{hariharan2017low} & ResNet-50 & 54.3 & 67.0 & 77.4 \\
MatchingNet \cite{vinyals2016matching} & ResNet-50 & 53.5 & 63.5 & 72.7 \\
ProtoNet \cite{snell2017prototypical} & ResNet-50 & 49.6 & 64.0 & 74.4 \\
PMN \cite{wang2018low} & ResNet-50 & 53.3 & 65.2 & 75.9 \\
\midrule
KTCH \cite{liu2019large} & ResNet-50 & 58.1 & 67.3 & 77.6 \\
KGTN \cite{chen2019knowledge} & ResNet-50 & 60.1 & 69.4 & 78.1 \\
\midrule
\textbf{SGM + Graph (Ours)} & ResNet-50 & \textbf{61.09} $\pm$ \textbf{0.37} & \textbf{70.35 $\pm$ 0.17} & \textbf{78.61 $\pm$ 0.19} \\
\bottomrule
\end{tabular}
\label{tab:imagenet-fs}
\end{table}

\subsection{Experiments on Synthetic Dataset}

To analyze the benefits of graph regularization, we devise a few-shot classification problem on a synthetic dataset. We first embed a balanced binary tree of height $h$ in $d$-dimensions using node2vec \cite{grover2016node2vec}. We set all leaf nodes as classes, and assign half as base and half as novel. For each task, we sample $k$ support and $q$ query examples from a Gaussian with mean centered at each class embedding and standard deviation $\sigma$. Given $k$ support examples, the task is to predict the correct class for query examples among novel classes. In these experiments, we set $d=4$, $h\in \{4,5,6,7\}$, $k \in \{1,2, ...,10\}$, $q=50$, and $\sigma \in \{0.1,0.2,0.4\}$. The baseline model is a linear classifier layer with cross-entropy loss, and we apply graph regularization to this baseline. We learn using SGD with learning rate 0.1 for 100 iterations.

\begin{figure}[h]
    \centering
    \includegraphics[width=.35\linewidth]{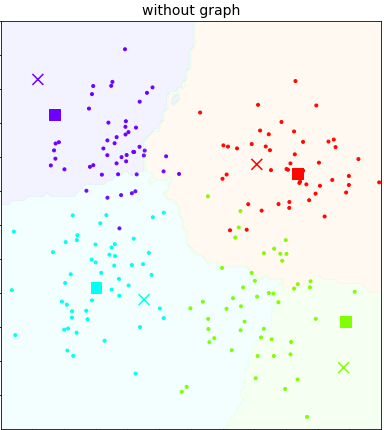}
    \hspace{0.5cm}
    \includegraphics[width=.35\linewidth]{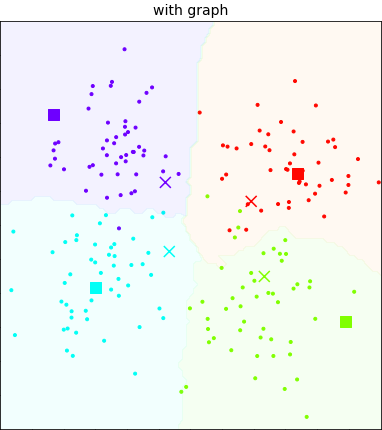}\\
     \caption{Synthetic experiment results. PCA visualization of learned classifiers for a single task without (left) and with graph regularization (right). Support examples are squares, query examples are dots, learned classifiers are crosses. Shaded regions show decision boundaries.}
    \label{fig:synthetic_viz}
\end{figure}

We first visualize the learned decision boundaries on identical tasks with and without graph regularization in \autoref{fig:synthetic_viz}. In this task, the sampled support examples are far away from the query examples, particularly for the purple and green classes. The baseline model learns poor decision boundaries, resulting in many misclassified query examples. In contrast, much fewer query examples are misclassified when graph regularization is applied. Intuitively, graph regularization helps more when the support set is further away from the sampled data points, and thus generalization is harder.

To measure the relationship between few-shot task difficulty and performance, we adopt the hardness metric proposed in \cite{dhillon2019baseline}. Intuitively, few-shot task hardness depends on the relative location of labeled and unlabeled examples. If labeled examples are close to the unlabeled examples of the same class, then learned classifiers will result in good decision boundaries and consequently accuracy will be high. 
Given a support set $\mathcal{D}_s$ and query set $\mathcal{D}_q$, the hardness $\Omega_{\phi}$ is defined as the average log-odds of a query example being classified incorrectly:

\begin{equation}
\Omega_{\phi} (\mathcal{D}_q;\mathcal{D}_s) = \frac{1}{N_q} \sum_{(x,y) \in \mathcal{D}_q} \log \frac{1 - p(y|x)}{p(y|x)}    
\end{equation}

where $p(\cdot| x_i)$ is a softmax distribution over $sim(x_i, p_j)=-\norm{x_i - p_j}^2_2$, the similarity scores between query examples $x_i$ and the means of the support examples $p_j$ from the $j^{th}$ class in $\mathcal{D}_s$.

\begin{figure}[h]
    \centering
    \includegraphics[width=.4\linewidth]{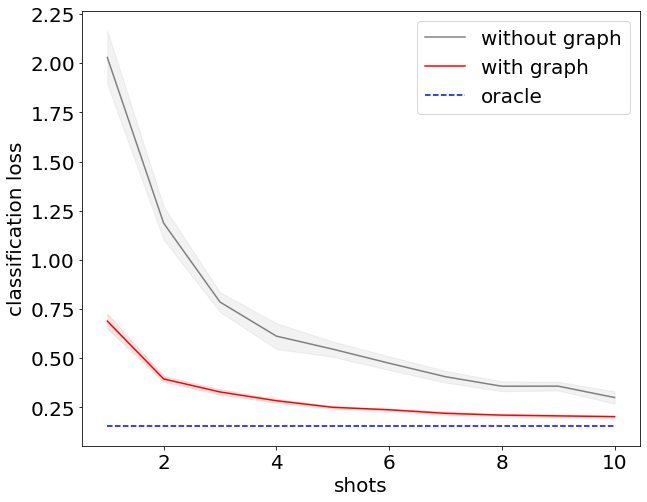}
    \includegraphics[width=.4\linewidth]{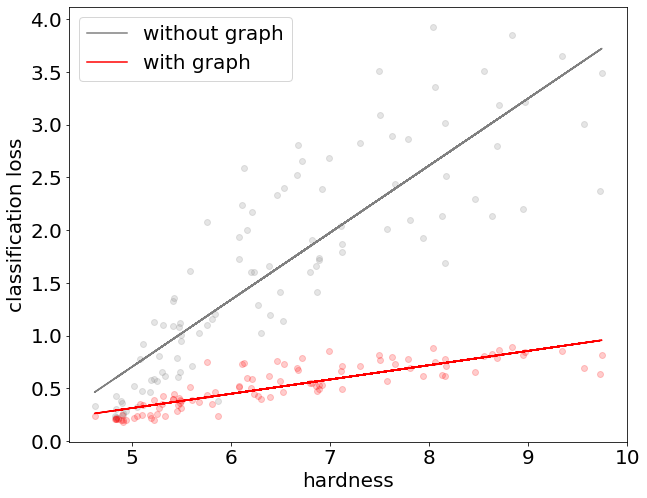}
    \caption{Quantified results of classification loss across shots (left) and task hardness metric (right). Each point is a sampled task. Red color denotes graph regularized method and gray method without graph regularization.}
    \label{fig:synthetic}
\end{figure}

We show average loss with shaded 95\% confidence intervals across shots in \autoref{fig:synthetic} (left), confirming our observations in real-world datasets that graph regularization improves the baseline model the most for tasks with lower shots. Furthermore, using our synthetic dataset, we artificially create more difficult few-shot tasks by increasing  $h$, tree heights, and increasing $\sigma$, the spread of sampled examples. We plot loss with respect to the proposed hardness metric of each task in \autoref{fig:synthetic} (right). The results  demonstrate that graph regularization achieves higher performance gains on more difficult tasks.

\section{Conclusion}

We have introduced a graph regularization method for incorporating label graph side-information into few-shot learning. Our approach is simple and effective, model-agnostic and boosts performance of a wide range of few-shot learners. We further showed that introduced graph regularization outperforms more complex state-of-the-art graph embedded models.

\begin{ack}
We thank Yueming Wang and Eli Pugh for discussions and providing feedback on our manuscript. We also gratefully acknowledge the support of DARPA under Nos. FA865018C7880 (ASED), N660011924033 (MCS); ARO under Nos. W911NF-16-1-0342 (MURI), W911NF-16-1-0171 (DURIP); NSF under Nos. OAC-1835598 (CINES), OAC-1934578 (HDR), CCF-1918940 (Expeditions), IIS-2030477 (RAPID); Stanford Data Science Initiative, Wu Tsai Neurosciences Institute, Chan Zuckerberg Biohub, Amazon, Boeing, JPMorgan Chase, Docomo, Hitachi, JD.com, KDDI, NVIDIA, Dell. J. L. is a Chan Zuckerberg Biohub investigator.
\end{ack}

\bibliography{refs}

\newpage
\begin{appendices}
\section{Problem Statement and Related Work}
\label{appendix:prob_setup}

\paragraph{Episodic Training} A common approach is to learn a few-shot model on $C_{train}$ in an episodic manner, so that training and evaluation conditions are matched \cite{triantafillou2019meta}. Note that training on support set examples during episode evaluation is distinct from training on $C_{train}$. Many metric based meta-learners and optimization based meta-learners use this training method, including Matching Networks \cite{vinyals2016matching}, Prototypical Networks \cite{snell2017prototypical}, Relation Networks \cite{sung2018learning}, and MAML \cite{finn2017model}.

\paragraph{Non-episodic Baselines} Inspired by the transfer learning paradigm of pre-training and fine-tuning, a natural non-episodic approach is to train a classifier on all examples in $C_{train}$ at once. After training, the final classification layer is removed, and this neural network is used as an embedding function $f$ that maps images $\textbf{x}_i$ to $x_i \in \RR$ feature representations, including those from novel classes. It then fine-tunes the final classifier layer using support set examples from the novel classes. The models are a function of the parameters of a softmax layer, $\theta \subset \RR^d$. The softmax layer is formulated as the similarity between image feature embeddings and the classifier parameters where $\theta_j$ is the parameters for the $j^{th}$ class, $sim$ is the cosine similarity function.
\begin{equation}
    p(y_i| x_i; \theta) = \frac{\exp(sim(x_i,  \theta_{y_i}))}{\sum_{y' \in \mathcal{Y}} \exp(sim(x_i, \theta_{y'}))}
\end{equation}

\subsection{Related work}

\paragraph{Few-Shot Learning} Canonical approaches to few-shot learning include memory-based \cite{gidaris2018dynamic, hariharan2017low, qiao2018few}, metric learning \cite{ravi2016optimization, vinyals2016matching, snell2017prototypical, sung2018learning}, and optimization-based methods \cite{finn2017model, rusu2018meta}. However, recent studies have shown that simple baseline learning techniques (\textit{i.e.}, simply training a backbone, then fine-tuning the output layer on a few labeled examples) outperform or match performance of many meta-learning methods \cite{chen2019closer, dhillon2019baseline}, prompting a closer look at the tasks \cite{triantafillou2019meta} and contexts in which meta-learning is helpful for few-shot learning \cite{raghu2019rapid, tian2020rethinking}. 

\paragraph{Few-Shot Learning with Graphs} Beyond the canonical few-shot literature, studies have explored learning GNNs over episodes as partially observed graphical models \cite{garcia2017few} and using GCNs to transfer knowledge of semantic labels and categorical relationships to unseen classes in zero-shot learning \cite{wang2018zero}. Recently, Chen et al. presented a knowledge graph transfer network (KGTN), which uses a Gated Graph Neural Network (GGNN) to propagate information from base categories to novel categories for few-shot learning \cite{chen2019knowledge}. Other works use domain knowledge graphs to provide task specific customization \cite{suo2020tadanet}, and propagate prototypes \cite{liu2019prototype, liu2019learning}. However, these models have highly complex architectures and consist of multiple sub-modules that all seem to impact performance. 

\section{Experimental Setup}
\label{appendix:exp_setup}

\subsection{Mini-ImageNet}
\paragraph{Dataset} The Mini-ImageNet dataset is a subset of ILSVRC-2012 \cite{deng2009imagenet}. The classes are randomly split into $64$, $16$ and $20$ classes for meta-training, meta-validation, and meta-testing respectively. Each class contains $600$ images. We use the commonly-used split proposed in \cite{vinyals2016matching}. 

\paragraph{Training details} We pre-train the feature extractor on $\mathcal{C}_\text{train}$ using the method proposed by \cite{mangla2020charting}. Activations in the penultimate layer are pre-computed and saved as feature embeddings of 640 dimensions to simplify the fine-tuning process. For an $N$-way $K$-shot problem, we sample $N$ novel classes per episode, sample $K$ support examples from those classes, and sample 15 query examples. During pre-training and meta-training stages, input images are normalized using the mean and standard-deviation computed on ILSVRC-2012. We apply standard data augmentation including random crop, left-right flip, and color jitter in both the training or meta-training stage. We use ResNet-18, ResNet-50 \cite{he2016deep}, and WRN-28-10 \cite{zagoruyko2016wide} for our backbone architectures. For pre-training WRN-28-10, we follow the original hyperparameters and training procedures for $\text{S2M2}_R$ \cite{mangla2020charting}. For meta-training ResNet-18, we follow the hyperparameters from \cite{chen2019closer}. At evaluation time, we choose hyperparameters based on performance on the meta-validation set. Some implementation details are adjusted for each method. Specifically, for ProtoNet and LEO, we include base examples during an additional adaptation step per class. We show that these alterations have a minimal contribution to performance in Appendix \ref{appendix:ablations}. 

\subsection{ImageNet-FS}
\paragraph{Dataset} In the ImageNet-FS benchmark task, the $1000$ ILSVRC-2012 categories are split into $389$ base categories and $611$ novel categories. From these, $193$ of the base categories and $300$ of the novel categories are used during cross-validation and the remaining $196$ base categories and $311$ novel categories are used for the final evaluation. Each base category has around $1,280$ training images and $50$ test images. 

\paragraph{Training details} We follow the procedure by \cite{hariharan2017low} to pre-train the ResNet-50 feature extractor, and adopt the Square Gradient Magnitude loss to regularize representation learning, which we scale by $0.005$. The model is trained using the SGD algorithm with a batch size of $256$, momentum of $0.9$ and weight decay of 0.0005. The learning rate is initialized as $0.1$ and is divided by $10$ for every 30 epochs. During fine-tuning, we train for $10,000$ iterations using the SGD algorithm with a batch size of 256, momentum of $0.9$, weight decay of 0.005, and learning rate of $0.01$. 

\subsection{Label Graph}

\paragraph{WordNet ontology} ImageNet comprises of $82,115$ synsets, which are based on the WordNet ontology. For both the Mini-ImageNet and ImageNet-FS experiments, we first choose the synsets corresponding to the output classes of each task -- $100$ for Mini-ImageNet and 1000 for ImageNet-FS. ImageNet provides IS-A relationships over the synsets, defining a DAG over the classes. We only consider the sub-graph consisting of the chosen classes and their ancestors. The classes are all leaves of the DAG. 

\paragraph{Training details} The hyperparameter settings used for the node2vec-based graph regularization objective are in line with values published in \cite{grover2016node2vec}. For all experiments, we set $p=1, q=1$ and temperature $T=2$. We set the batch size to $128$ for Mini-ImageNet and $256$ for ImageNet-FS. Empirically, we find that setting the regularization $\lambda$ scaling higher for lower shots results in better performance, and set $\lambda=5,3,1$ for 1-,2-, and 5-shot tasks respectively.

\section{Ablations}
\label{appendix:ablations}

\subsection{Mini-ImageNet Ablations}

\subsubsection{Model re-implementations with adaptation}
For episodically-evaluated few-shot models, it is common practice to disregard base classes during evaluation. To implement graph regularization, we include both base and novel classes during test time and perform a further adaptation step per task. We show that the boost in performance is not due to these modifications. 

\begin{table}[H]
\caption{Validation of baseline model modifications.}
\centering
\begin{tabular}{p{6cm}p{2cm}p{2cm}p{2cm}}
\toprule
\bf Model & \bf Backbone & \bf 1-shot & \bf 5-shot \\
\midrule
ProtoNet & ResNet-18 & 54.16 $\pm$ 0.82 & 73.68 $\pm$ 0.65 \\
ProtoNet (adaptation)$^{\dag}$ & ResNet-18 & 54.86 $\pm$ 0.73 & 74.14 $\pm$ 0.50 \\
\textbf{ProtoNet (adaptation) + Graph (Ours)} & ResNet-18 & \textbf{55.47 $\pm$ 0.73} & \textbf{74.56 $\pm$ 0.49} \\
\midrule
LEO$^{\dag}$ & WRN 28-10 & 58.22 $\pm$ 0.09 & 74.46 $\pm$ 0.19 \\
LEO (adaptation) & WRN 28-10 & 57.85 $\pm$ 0.20 & 74.25 $\pm$ 0.17 \\
\textbf{LEO (adaptation) + Graph (Ours)} & WRN 28-10 & \textbf{60.93 $\pm$ 0.19} & \textbf{76.33 $\pm$ 0.17} \\
\bottomrule
\end{tabular}
\end{table}

\subsubsection{Finding good parameter initializations for novel classes}
Recent works have shown that good parameter initialization is important for few-shot adaptations \cite{raghu2019rapid}. For example, Dhillion et al. \cite{dhillon2019baseline} showed that initializing novel classifiers with the mean of the support set improves few-shot performance. 

Here, we explore various methods of incorporating graph relations to improve parameter initialization for novel classes. We compare our proposed method with simpler methods to show that the our graph regularization method is boosting performance in a non-trivial manner. For each method, we keep the adaptation procedure the same, namely, the fine-tuning procedure described by Baseline++ \cite{chen2019closer}. 

We then vary parameter initialization using the following methods: (A) random initialization, (B) initializing novel classes with the weights of the closest training class in graph distance in the knowledge graph, (C) our method.

\begin{table}[H]
\centering
\caption{Mini-Imagenet with different parameter initialization methods (in \% measured over 600 evaluation iterations).}
\begin{tabular}{p{5cm}p{2cm}p{2cm}p{2cm}} \toprule
\bf Model & \bf Backbone & \bf 1-shot & \bf 5-shot \\
\midrule
$\text{S2M2}_R$ + Init A \citep{mangla2020charting} & WRN 28-10 & 64.93 $\pm$ 0.18 & 83.18 $\pm$ 0.11 \\
$\text{S2M2}_R$ + Init B & WRN 28-10 & 65.50 $\pm$ 0.81 &  83.32 $\pm$ 0.57 \\
\textbf{$\text{S2M2}_R$ + Init C} & WRN 28-10 & \textbf{66.93 $\pm$ 0.65} &  \textbf{83.35 $\pm$ 0.53} \\
\bottomrule
\end{tabular}
\end{table}

\subsection{ImageNet-FS Ablations}

Here, we justify our model design decisions by considering alternatives. We first 
probe the benefits of using random walk neighborhoods by defining $N(y)$ as only nodes that have direct edges with $y$ (``child-parent loss''). We try separately learning label graph embeddings, and passing the information to the classifier layer via ``soft target'' classification loss (``Independent graph w/ soft targets''). Results show that computing the graph loss directly on the classifier parameters is important for performance. Finally, we show that the quality of the label graph affects performance by removing layers of internal nodes of the WordNet hierarchy, starting from the bottom-most nodes (``Remove last 5, 10 layers'').

\begin{table}[H]
\centering
\caption{Imagenet-FS ablations. Experiment setups, in order from the top: our proposed method, using only child-parent edges, independently learning graph embeddings, removing 5 layers of the ImageNet hierarchy, and removing 10 layers of the ImageNet hierarchy.}
\begin{tabular}{p{5cm}p{2cm}p{2cm}}
\toprule
\bf Ablation & \textbf{1-shot}  \\
\midrule
\textbf{Ours} & \textbf{61.09} \\
Child-parent loss & 56.78 \\
Independent graph w/ soft targets & 56.22  \\
Remove last 5 layers & 57.80 \\
Remove last 10 layers & 54.86 \\
\bottomrule
\end{tabular}
\end{table}

\end{appendices}
\end{document}